\documentclass[letterpaper, 10pt, journal]{IEEEtran}
\usepackage{silence}
\IEEEoverridecommandlockouts
\usepackage[T1]{fontenc}
\usepackage[latin1]{inputenc}
\usepackage{balance}
\usepackage{cite}
\usepackage{graphicx}
\usepackage{stfloats}
\usepackage{placeins}
\usepackage{url}
\usepackage{colortbl}
\usepackage{diagbox}
\usepackage{multirow}
\usepackage{enumerate}
\usepackage{siunitx}

\usepackage{amsmath,amssymb,amsfonts}
\usepackage{amsthm}
\usepackage{bm}
\usepackage{textcomp}
\usepackage[dvipsnames]{xcolor}
\usepackage{tikz}
\usepackage[export]{adjustbox}
\usepackage{subcaption}
\usepackage{booktabs}
\usepackage{cuted}
\usepackage[labelfont=bf,font=footnotesize]{caption}
\usepackage{hyperref}
\hypersetup{colorlinks=true, linkcolor=black, citecolor=black, urlcolor=blue}
\usepackage{cleveref}

\newcommand{\figdir}{figures}
\definecolor{catMemory}{HTML}{999999}
\definecolor{catHistory}{HTML}{a6cee3}
\definecolor{catRecurrent}{HTML}{b2df8a}
\definecolor{catAttention}{HTML}{fdbf6f}
\definecolor{catNoModel}{HTML}{bbbbbb}

\newcommand{\marker}[1]{\begin{tikzpicture}[baseline=-0.5ex]
    \fill[#1] (0,0) circle (3pt);
  \end{tikzpicture}}

\makeatletter
\let\NAT@parse\undefined
\makeatother

\title{Wake Up to the Past: Using Memory \\ to Model Fluid Wake Effects on Robots}

\author{Luca Vendruscolo, Eduardo Sebasti\'{a}n, Amanda Prorok, and Ajay Shankar%
\thanks{The authors are with the Department of Computer Science and Technology, University of Cambridge, UK. {\tt\footnotesize \{lv407, es2121, asp45, as3233\}@cst.cam.ac.uk}}%
\thanks{This work has been submitted to the IEEE for possible publication. Copyright may be transferred without notice, after which this version may no longer be accessible.}}

\begin{document}
\maketitle

\tikzset{
  panel label/.style={fill=white, fill opacity=0.9, text opacity=1,
                      draw=black!35, line width=0.3pt, rounded corners=1.5pt,
                      inner xsep=3pt, inner ysep=2pt, font=\footnotesize},
  lab n/.style={anchor=north west,
                at={([xshift=3pt,yshift=-3pt]tile.north west)}},
  lab s/.style={anchor=south west,
                at={([xshift=3pt,yshift=3pt]tile.south west)}},
}

\newlength{\tilewidth}
\newsavebox{\tilebox}
\newcommand{\teasercol}[5][0pt]{\begin{minipage}[t]{0.32\textwidth}\centering
    \setlength{\tilewidth}{\linewidth}\sbox{\tilebox}{\includegraphics[width=\tilewidth,frame]{\figdir/tile_#2.jpg}}\begin{tikzpicture}
      \path[use as bounding box]
        (0,-\the\dp\tilebox) rectangle (\the\wd\tilebox,\the\ht\tilebox);
      \node[inner sep=0pt, outer sep=0pt, anchor=south west] (tile)
        at (0,-\the\dp\tilebox) {\usebox{\tilebox}};
      \node[panel label, lab #5] {#4};
    \end{tikzpicture}\\[3pt]
    \includegraphics[width=\linewidth]{\figdir/ral_teaser_#3.pdf}\par\vspace{#1}\end{minipage}}

\begin{strip}
    \vspace{-47.5pt}
    \centering
    \makebox[\textwidth]{\begin{tabular}{@{}c@{\hspace{0.02\textwidth}}c@{\hspace{0.02\textwidth}}c@{}}
    \teasercol[3pt]{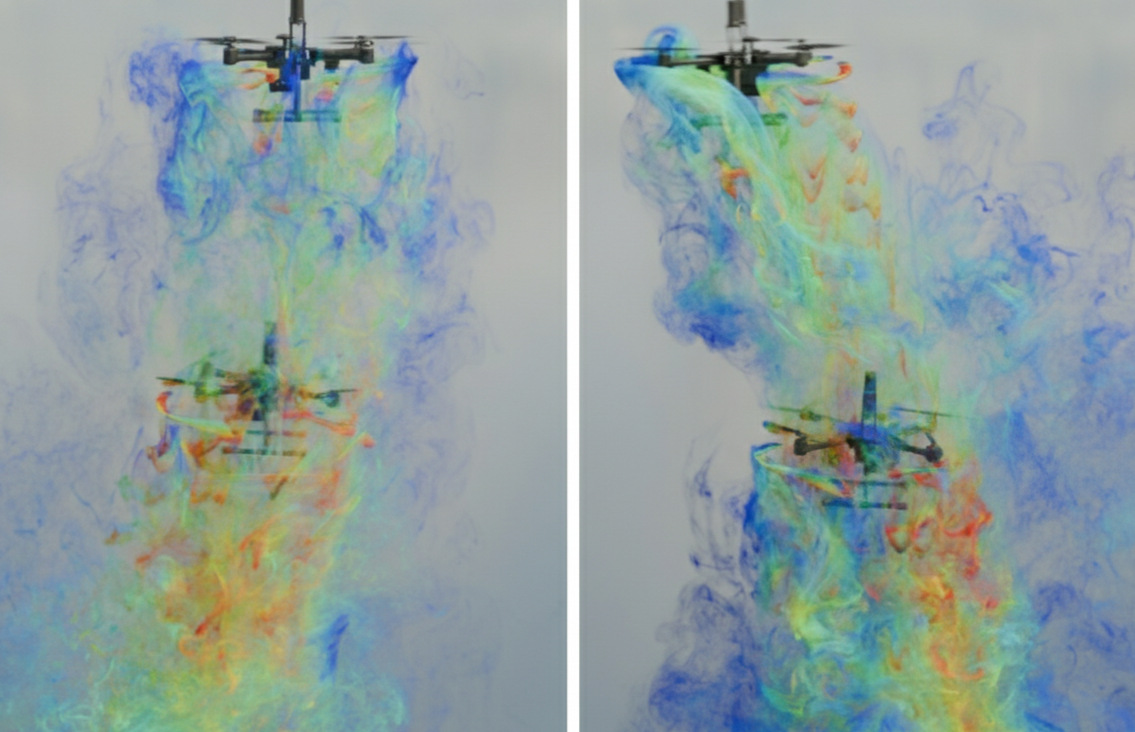}{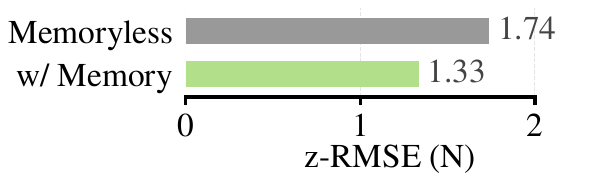}{(a) FlareDW dataset}{s}
    & \teasercol[3pt]{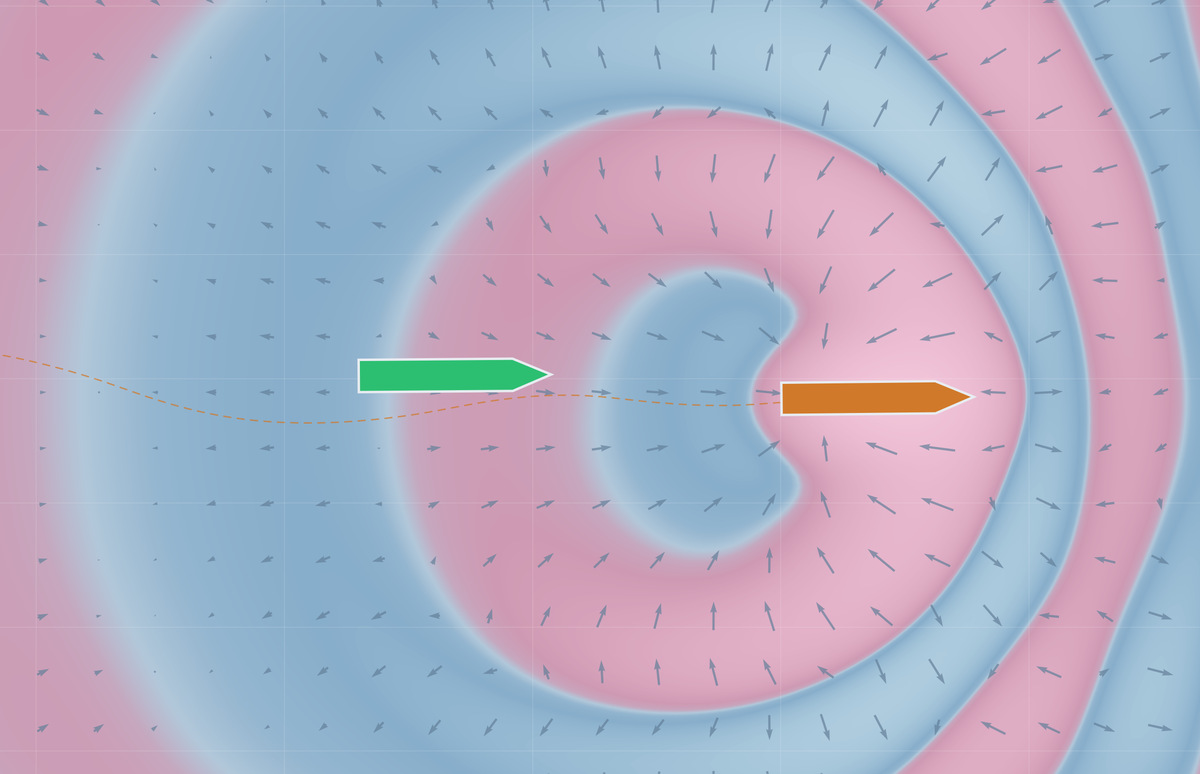}{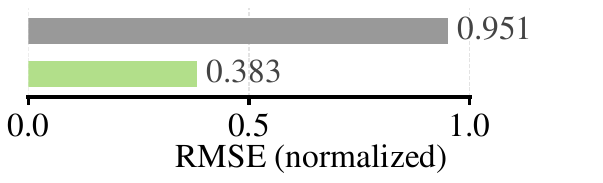}{(b) Ship--ship interactions}{s}
    & \teasercol[3pt]{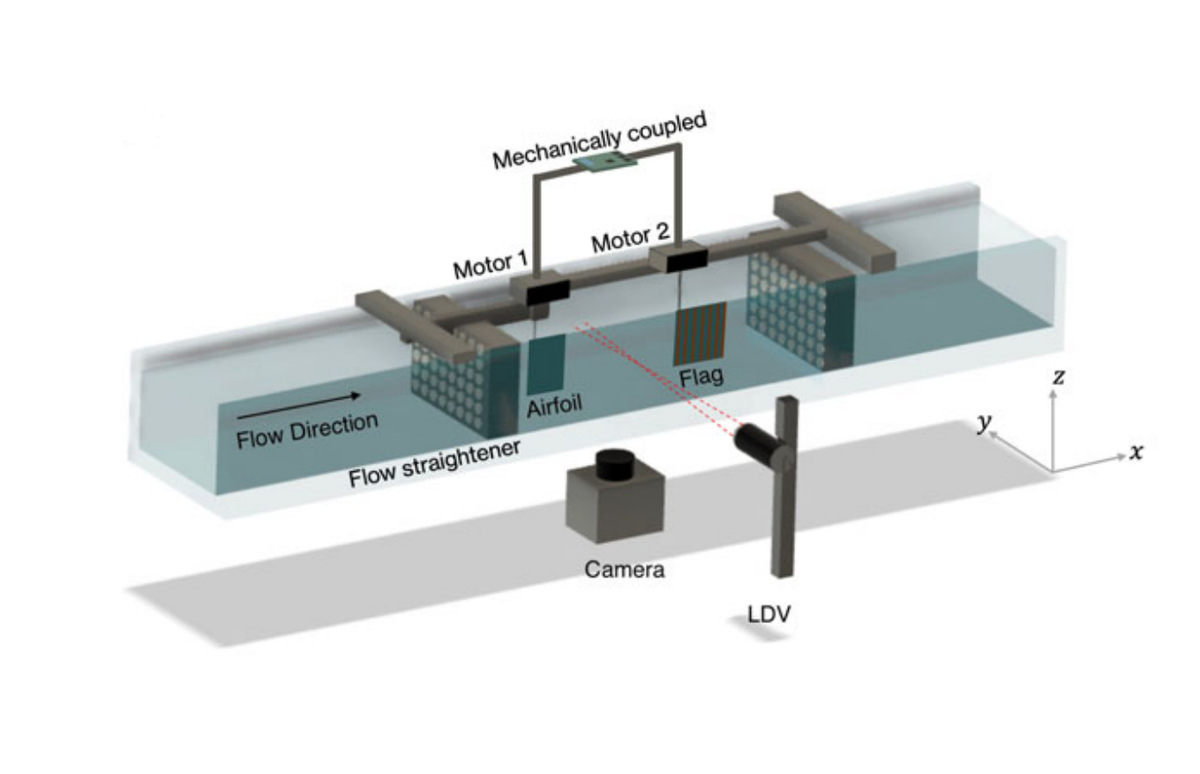}{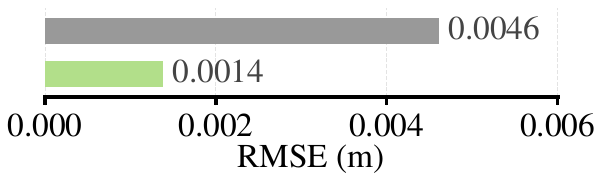}{(c) Fish schooling}{s}
    \end{tabular}}\\
    \makebox[\textwidth]{\begin{tabular}{@{}c@{\hspace{0.02\textwidth}}c@{\hspace{0.02\textwidth}}c@{}}
    \teasercol{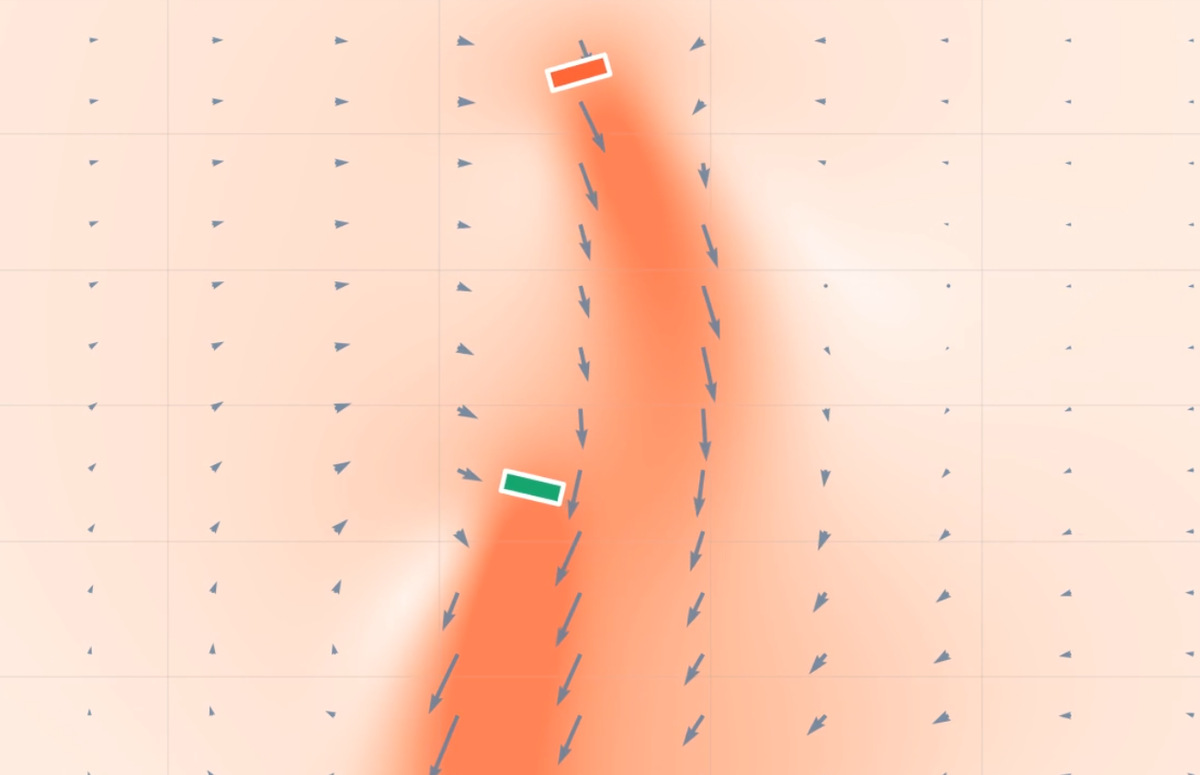}{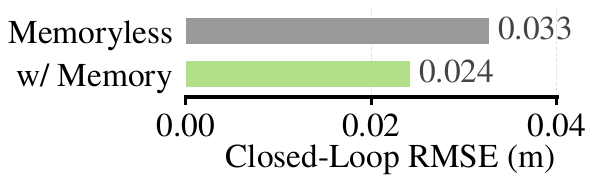}{(d) 2D Navier--Stokes simulator}{s}
    & \teasercol{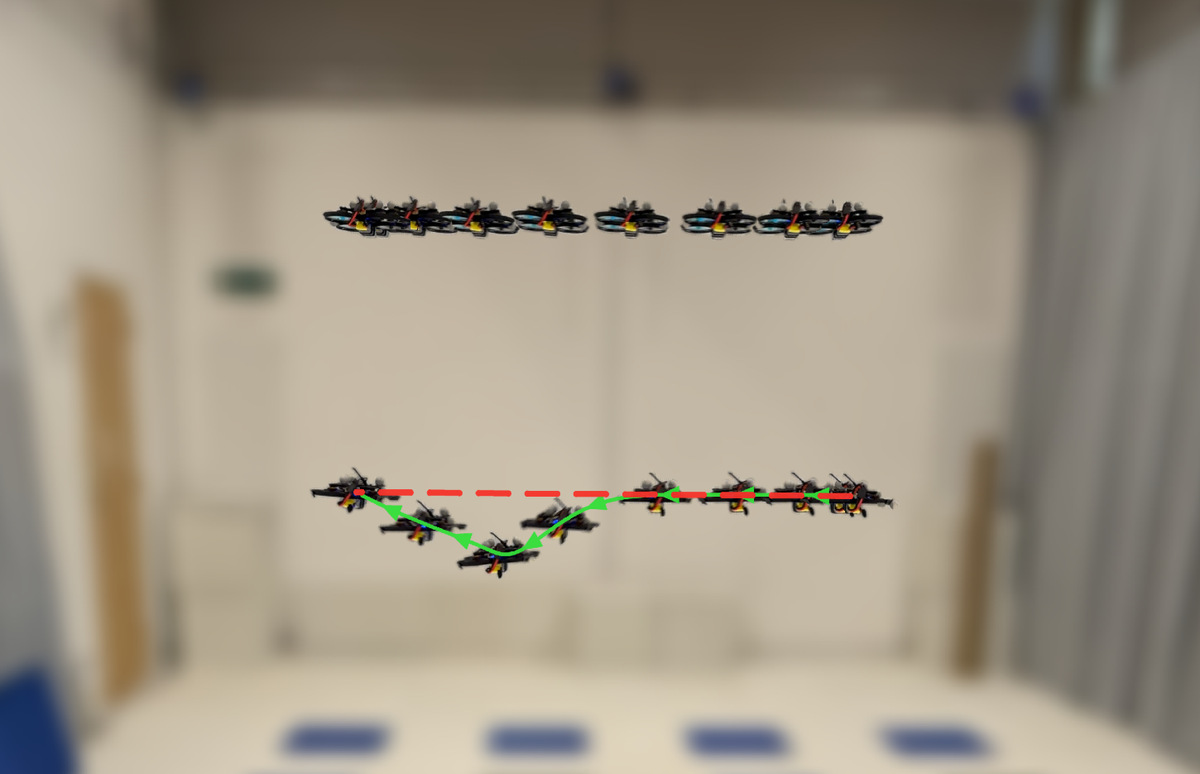}{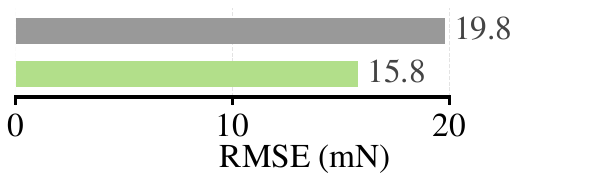}{(e) Free-flying quadrotors}{s}
    & \teasercol{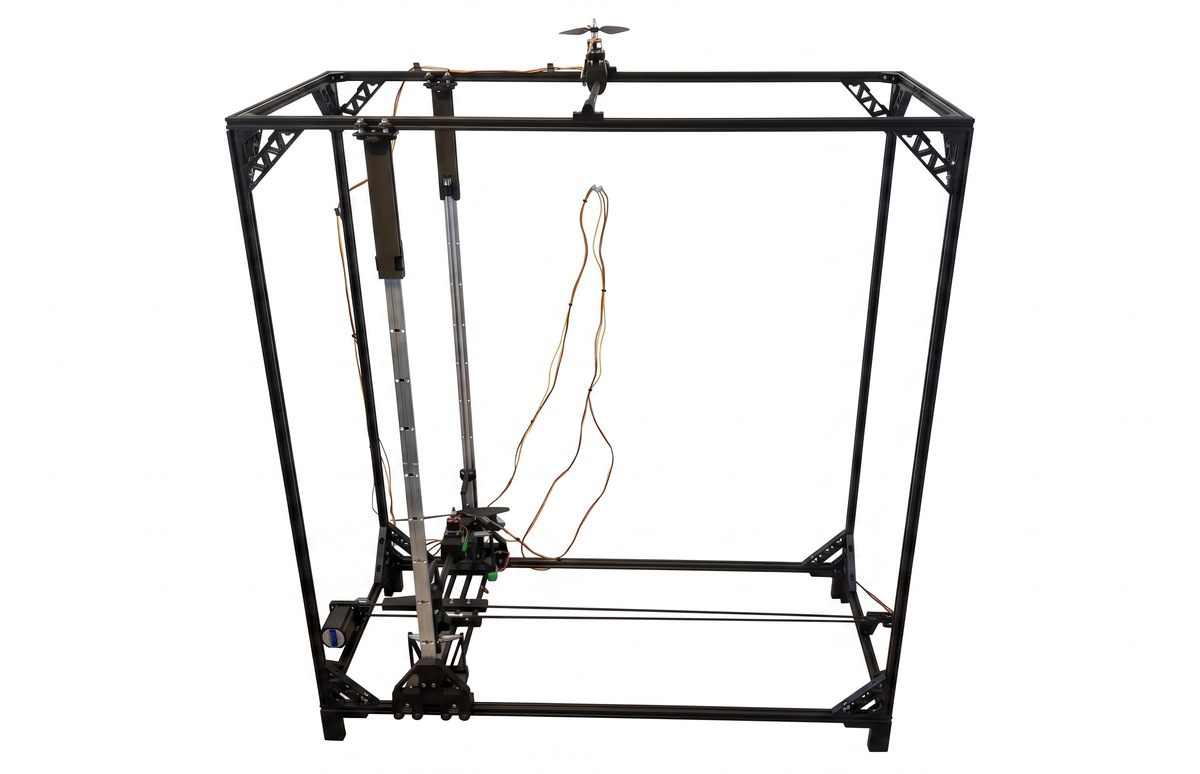}{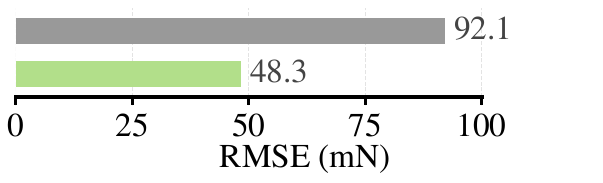}{(f) Monocopter gantry}{s}
    \end{tabular}}
    \vspace{4pt}
    \captionof{figure}{
    \textit{Comparing models with and without memory:}
    we study the influence of modeling temporal context in the prediction of fluid wake effects in six aerial and aquatic domains.
    In dynamic regimes, data-driven predictors benefit from history, which can account for the transport delay in the medium.
    Memory lowers the error in every domain: \SI{23}{\percent} (\textbf{a}), \SI{60}{\percent} (\textbf{b}), \SI{70}{\percent} (\textbf{c}), \SI{26}{\percent} (\textbf{d}), \SI{20}{\percent} (\textbf{e}), \SI{48}{\percent} (\textbf{f}).
    Illustrations in (a) and (c) are reproduced from \cite{kharitenko2025spatiotemporal} and \cite{das2025fish}.}
    \label{fig:teaser}
\end{strip}

\begin{abstract}
Autonomous aerial and aquatic robots that move by displacing their medium, such as multicopters and underwater vehicles, produce wake effects that disturb adjacent robots.
Wake effects are hard to model and predict due to the chaotic spatio-temporal dynamics of the fluid, entangled with the geometry and motion of the robots. Data-driven approaches using neural networks typically learn a memory-less function that maps the current states of the two robots to a force observed by the ``sufferer'' robot.
Such models often perform poorly, since the wake effect has a finite propagation time and the disturbance on the sufferer is a function of relative states in the past.
In this work, we study the temporal properties that a wake-effect predictor must capture to model such fluid-mediated interactions accurately.
We explore seven data-driven models designed to capture the spatio-temporal evolution of fluid wake effects in six different media, each compared against a state-of-the-art memoryless baseline.
Across all domains, we show that models with temporal context noticeably outperform the memoryless baseline while maintaining a reasonable computational footprint, even in closed-loop online deployments.
Web: \texttt{https://wake-up-to-the-past.vercel.app/}
\end{abstract}


\section{Introduction}\label{sec:intro}

\IEEEPARstart{R}{obots} that fly or swim generate momentum by pushing against the air or water around them.
This interaction perturbs the medium and produces fluid wake effects---such as the downwash from a multicopter or the trailing vortex of an autonomous underwater vehicle---that act as significant disturbances for adjacent robots~\cite{gielis2023modeling}.
Accurately modeling and compensating for these effects is thus crucial for safe and reliable operation of a fleet in close proximity~\cite{shankar2023docking,li2021energy}.
The disturbance can in principle be measured with dedicated sensors~\cite{davis2017direct, krieg2019distributed, nelson2020hydrodynamic}, but these require hardware, payload and calibration that are expensive and complex to integrate. A predictor instead infers the disturbance from the robots' relative motion.
While humans can manually correct for these forces during specific maneuvers like mid-flight refueling or ship docking, robot deployments with unknown motion patterns require predictors that are accurate and computationally lean.

The primary challenge in modeling these effects is the chaotic spatio-temporal dynamics of the fluid, which are linked to the geometry and motion of the robots~\cite{brouzet2025undulatory,gielis2023modeling}.
Robots perturb the medium in many ways (a fish-like caudal fin, a propeller, a paddle), and each sheds a different chaotic wake, which is part of why we study several media. Analytical models are too simple to capture such wakes~\cite{leishman2006principles, jain2019modeling}, while computational fluid dynamics can take hours to simulate minutes of motion~\cite{venturadiaz2018highfidelity, fu2025assessment}. Neither is practical for disturbance compensation in a control loop, which is the ultimate goal of a predictor.
Existing observation schemes typically include current velocity and relative velocity as variables that inform the evolution of the interaction between the robots~\cite{smith2023so,kiran2025influence}. However, instantaneous information is insufficient for capturing the \textit{mechanical} aspects of the medium, in particular, transport delays in the propagation of wake effects.
Because wake effects have a finite propagation time, the disturbance currently affecting a sufferer robot is a function of the ``wake source'' robot's relative state at a point in the past.
Without temporal context, the sufferer is unable to correlate the current relative state with the current disturbance.
Consequently, memory-less models perform poorly in agile scenarios.
As an illustrative example, consider two multirotors whose instantaneous observation is identical, one having just reached that velocity, the other having cruised at it throughout. The air beneath the first still carries momentum shed while it was stationary, so the wake columns differ, yet a model of the current state returns the same prediction for both.
In spite of this dependency,  analyses do not incorporate history and recurrence in wake-effect prediction models for robots.

The goal of this paper is to fill this gap and shed light on which properties of a data-driven predictor matter for accurate modeling of wake effects at a low computational cost, such that the predictor can be deployed in real-time feedback.
Since using history to overcome partial observability is well established in other domains~\cite{ji2025trainable,brunton2017chaos,saviolo2022physics}, we test how it transfers to wake prediction, and what that implies for choosing a model. 
Specifically, our contributions are:

\begin{itemize}
    \item An empirical study of wake effects across seven data-driven architectures, each compared against a state-of-the-art memoryless baseline, and six domains. The architectures range from simple history-based MLPs to advanced sequential models such as Mamba~\cite{gu2023mamba}, to determine which temporal mechanisms best capture fluid dynamics. The domains span CFD and experimental datasets with aerial and aquatic robots, a shallow-water ship-convoy simulator, and a 2D Navier--Stokes downwash simulator with prediction-in-the-loop flight. \gdef\introtakeaway{The results reveal where the gain comes from and when each model is worth its cost, which grounds our model-selection guidance.}\introtakeaway
    \item Real-world validation on a custom monocopter gantry and a pair of free-flying lightweight quadrotors, evaluated on diverse flights and scenarios, with closed-loop compensation across different randomized flight regimes.
\end{itemize}

\noindent\textbf{Related Work.}\label{sec:related}
Modeling inter-robot wake interaction effects has seen substantial interest in recent years, particularly in the aerial domain.
Multirotor downwash, for instance, can exhibit high variability in turbulence depending on the flight regime~\cite{zhang2020numerical}, and has non-trivial effects on neighboring robots~\cite{gielis2023modeling}.
These effects have been studied extensively using numerical models~\cite{kharitenko2025spatiotemporal,zhang2020numerical} as well as some empirical validation~\cite{jain2019modeling, kiran2025influence}.
\looseness=-1 Computational fluid modeling has also been applied to aquatic surface and sub-surface vehicles~\cite{tavakoli2022wake}.

Of particular interest is the ability to incorporate a given model into a real-time feedback control and planning framework, i.e., the ability to \textit{reject} the disturbance.
To this end, recent work has modeled these effects on multirotors as forces and torques acting on a sufferer, and has developed methods to predict them using numerical models~\cite{kiran2025influence}, and neural networks with differential~\cite{li2023nonlinear, shi2020neural, shi2021neural,chee2025flying} and geometric priors~\cite{smith2023so, kharitenko2025spatiotemporal}. In aquatic domains, distributed networks of sensors have been integrated with hydrodynamic models to predict the forces suffered by underwater vehicles \cite{krieg2019distributed, nelson2020hydrodynamic}, and early research considered wake-effect predictors for control compensation in aquatic manipulation \cite{levesque1994dynamic}.
From a biological perspective, studies on fish schooling~\cite{li2021energy} have shown the importance of wake-effect prediction not only for stability, but also for energy efficiency in robot motion. Inspired by them, current methods for aquatic robots explore computational fluid simulators and pressure-based analytical methods for soft robots that mimic the animals' mechanical structure \cite{xu2013fish}, and similar approaches are found for bio-inspired flapping-wing robots \cite{armanini2017onboard}.

While these data-driven approaches have shown impressive results, we observe that they tend to perform well in slow-moving, near-steady-state, laminar flow, which admits modeling the effect at a given time as a function of the relative state \textit{at that time}. In highly dynamic settings, the impact on a sufferer instead results from some relative state in the past, and the instantaneous state is a poor approximation.
Outside wake prediction, time-delay identification includes networks that learn their delays from data~\cite{ji2025trainable} and delay-coordinate models of chaos~\cite{brunton2017chaos}, and physics-informed temporal modeling has improved a single quadrotor's dynamics model~\cite{saviolo2022physics}. Our aim is different: rather than identify a delay, we show empirically that generic temporal models learn behavior consistent with physical transport. 

\begin{table*}[htbp]
\centering
\caption{Summary of model architectures. Parameter counts are medians across domains. Hyperparameters and sizes are re-tuned per domain.}
\label{tab:architectures}
\renewcommand{\arraystretch}{0.95}
\begin{tabular}{@{}lllll@{}}
\toprule
\textbf{Architecture} & \textbf{Parameters} & \textbf{Temporal Mechanism} & \textbf{Model-Specific Tuned Hyperparameters} & \textbf{Category} \\ \midrule

Agile MLP \cite{kharitenko2025spatiotemporal} & 27K & Instantaneous input features & --- & \marker{catMemory}  Memoryless
\\
History MLP \cite{bourlard1988links} & 21K & Flattened window of past snapshots & \#snapshots & \marker{catHistory} Explicit History
\\
Delay Embedding \cite{yang2019wide} & 25K & Learned Gaussian kernel over history & $\sigma_0$, selector dimensions & \marker{catHistory} Explicit History
\\
GRU \cite{chung2014empirical} & 81K & Learned hidden state & projection dimensions & \marker{catRecurrent} Recurrent
\\
TCN \cite{bai2018tcn} & 28K & Causal dilated 1D convolutions & kernel size & \marker{catRecurrent} Recurrent
\\
Mamba \cite{gu2023mamba} & 36K & Selective state-space model & $d_\text{model}$, $d_\text{state}$, convolution width, expansion &  \marker{catRecurrent} Recurrent
\\
RC (ESN) \cite{jaeger2002adaptive} & 1K* & Hidden state from fixed random reservoir & reservoir size, spectral radius, leak rate & \marker{catRecurrent} Recurrent
\\
Cross-Attention \cite{vaswani2017attention} & 34K & Weighted combination of processed tokens & \#heads, embed size, key spacing, temperature & \marker{catAttention}  Attention-based
\\ \bottomrule
\multicolumn{5}{@{}l}{\footnotesize *RC (ESN) uses additional 3,568K frozen weights in the reservoir.}
\end{tabular}
\end{table*}

\section{Preliminaries}\label{sec:problem}

Consider two robots, a wake source and a sufferer, moving in a fluid medium.
The wake source's actuation at a past time $t-\Delta t$ perturbs the medium, creating a wake effect (e.g., downwash or trailing vortices).
This perturbation propagates through the fluid at a finite speed and reaches the sufferer at the present time $t$, which we model as a disturbance force \mbox{$F_{db} \in \mathbb{R}^3$}.
The objective of a wake-effect predictor is to learn a mapping $\Phi$ that \textit{predicts} the instantaneous disturbance force $F_{db}(t)$ experienced by the sufferer. We assume that the wake source and sufferer can cooperate, and thus, wake source can communicate unobservable information (such as its thrust vector) to the sufferer when necessary.

For training we assume a measure of $F_{db}$ is available, recovered from the sufferer's tracked motion and its known actuation commands, as in prior work~\cite{smith2023so}, or provided directly by a simulator.
We collect all relevant data into a time-indexed dataset of the full state of the two robots and $F_{db}$, from which an observation vector $o(t) \in \mathbb{R}^d$ is built and then preprocessed or concatenated with past data, depending on the architecture, to form an input vector $O(t)$.
We train a predictor over this dataset using a supervised loss to learn the mapping $\Phi$ with a neural network
$
\hat{F}_{db}(t) = f(O(t), \theta),
$
where $\theta$ are learnable parameters and $\hat{\cdot}$ marks a prediction.

The goal of this work is to study the properties that $O$ and $f$ must satisfy in order to make accurate predictions of $F_{db}$.
In particular, we test the key hypothesis that encoding temporal context and modeling propagation delay $\Delta t$ are essential in highly agile and dynamic settings. The specific observation vector $O$ varies across domains, but generally encodes relative position and velocity between robots, actuation inputs (e.g., thrust or rudder commands), and domain-specific kinematic quantities. Regarding the delay $\Delta t$, we remark that it is a property of the medium, and in every domain it follows from the flow. For a rotor, momentum theory gives an induced velocity $w \propto \sqrt{T/(\rho A)}$ for thrust $T$ over disc area $A$~\cite{leishman2006principles}, so a wake column crossing a vertical separation $\Delta z$ arrives after ${\sim}\Delta z/w$. The lag grows with separation, shortens as the source works harder, and lengthens in a denser medium. In shallow water it is set instead by the non-dispersive long-wave speed $\sqrt{gh}$, and for the fish by advection with the tunnel flow. Nonetheless, due to the chaotic nature of the medium, it is hard to analytically model $\Delta t$ accurately, which motivates testing the hypothesis on learning a representation of such delay.

\section{Model Architectures}\label{sec:architectures}

We investigate a diverse range of data-driven predictors categorized by their mechanism for handling temporal context, described below and summarized in Table \ref{tab:architectures}. The goal is to build a common framework to make results comparable across domains. The hyperparameters of every model are tuned for each domain through Bayesian optimization using Optuna \cite{akiba2019optuna}, keeping the trainable parameters below $100$K so the models are computationally lean and, potentially, implementable on board a robot.
Each search runs 50 trials of up to 120 epochs, and the best configuration is retrained for up to 600 epochs, with early stopping in both stages. 

\subsection{Memoryless Baseline}

Agile MLP \cite{kharitenko2025spatiotemporal} is a memory-less feedforward network that maps the $d-$dimensional input equivariant feature vector
directly to a 3D force prediction in the equivariant frame. It assumes the fluid wake is a function of the instantaneous relative state (geometry, velocity, thrust).
It has recently been shown that this parameterization outperforms other existing memory-less models in pair-wise downwash prediction in quadrotors \cite{li2023nonlinear, smith2023so, shi2020neural, shi2021neural},
justifying it as a primary baseline.

\subsection{Explicit History Models}

These models explicitly provide a history of past snapshots to capture temporal changes without recurrence.

\subsubsection{History MLP}

The History MLP concatenates multiple past snapshots, evenly spaced over a tunable time window, into a single flattened input vector. By observing a sliding window of recent history, the model can infer which wake patterns generated in the past are arriving at the sufferer now.

\subsubsection{Delay Embedding}

The Delay Embedding model learns to pick a snapshot at time $\tau$ rather than processing a full window of snapshots. Since the wake travels at a finite speed, a small MLP maps the current observation to a delay $\tau$, the center of a Gaussian kernel with one learned width $\sigma$ that averages the history into a single delayed snapshot. That snapshot is fed to an MLP along with the current observation.

\subsection{Recurrent Models}

\looseness=-1 These models maintain an internal hidden state to summarize the history of observations, rather than holding an explicit time-series in the input.

\subsubsection{Gated Recurrent Unit (GRU)}
GRUs process observations sequentially, maintaining a compressed summary of the history in a hidden state. Learned gates (reset and update) allow the model to selectively remember stable wake patterns or rapidly update the state during agile maneuvers.
\subsubsection{Temporal Convolutional Network}
The Temporal Convolutional Network (TCN) processes trajectories using causal dilated 1D convolutions. With exponentially increasing dilations, the TCN builds a receptive field well beyond the typical transport delays, detecting patterns at multiple timescales.

\subsubsection{Mamba}
Mamba is a selective state-space model whose recurrence parameters depend on the input. It lets the model retain or discard information at every timestep, a learned counterpart to fixed-recurrence systems.

\subsubsection{Reservoir Computing with Echo State Networks}
Reservoir Computing with Echo State Networks (RC-ESN) uses a large, fixed, randomly wired recurrent reservoir and only learns the readout layer. Inspired by cognitive models of the brain, a high-dimensional random dynamical system naturally creates ``echoes'' of past inputs, allowing a linear readout to pick the most predictive temporal features. The reservoir's frozen weights still contribute to computation.

\subsection{Cross-Attention}
The Cross-Attention model allows the sufferer to selectively query relevant moments from the wake source's past trajectory. In practice, the network is based on multi-head cross-attention with fixed sinusoidal positional encodings.

\section{Evaluations}\label{sec:evals}

\begin{figure*}
    \centering
    \includegraphics[width=\textwidth]{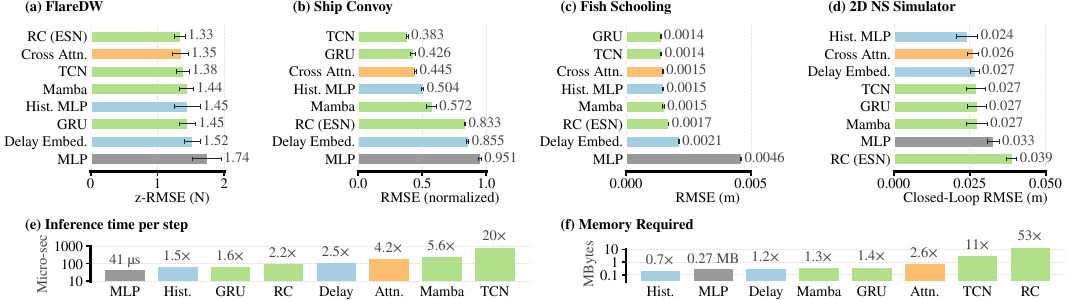}
    \caption{Performance and inference cost of each method on the four main domains. (\textbf{a}) A CFD dataset for two quadrotors~\cite{kharitenko2025spatiotemporal}. (\textbf{b}) Our shallow-water ship-convoy simulator, whose wave field is generated by a moving pressure patch~\cite{david2017shipwaves}. (\textbf{c}) A dataset that captures hydrodynamic interactions between two fish swimming in line~\cite{das2025fish}. (\textbf{d}) Closed-loop tracking on our 2D Navier--Stokes downwash simulator. Against the memoryless MLP, the best model with memory cuts the error by \SI{23}{\percent} (\textbf{a}), \SI{60}{\percent} (\textbf{b}), \SI{70}{\percent} (\textbf{c}) and \SI{26}{\percent} (\textbf{d}). Error bars show $\pm 1$ standard deviation across seeds. (\textbf{e}, \textbf{f}) Time per prediction on one AMD Ryzen 7 7800X3D core (median over domains) and memory of the largest configuration (with RC's frozen weights), as multiples of the memoryless MLP's.}
    \label{fig:bar_plots}
\end{figure*}

\looseness=-1 We now evaluate each of the temporal-context models and the memory-less baselines on six domains:
a CFD dataset for two quadrotors~\cite{kharitenko2025spatiotemporal},
a custom wave-field simulator of a two-ship convoy in shallow water, a two-fish schooling dataset \cite{das2025fish}, a custom 2D Navier--Stokes downwash simulator, a pair of free-flying lightweight quadrotors, and a tethered monocopter gantry.
In the first three domains and the gantry we study prediction accuracy; in the 2D simulator and on the free-flying quadrotors we also consider closed-loop tracking with the predictions in the control loop.
We split the data into episodes and assign each to the training, validation or test set (\Cref{tab:protocol}). Each episode starts with an unscored warm-up period, and recurrent states start at zero. We set the warm-up by hand to be longer than the wake delay, so the models can build up their history. The labels are smoothed offline using future samples, but the models only see past and present inputs. 

\begin{table}[t]
\centering
\caption{Evaluation protocol per domain. Length is that of one episode. A seed sets the initialization and, except for the fish, redraws the split (the test set is fixed for the ship and 2D closed loop).}
\label{tab:protocol}
\setlength{\tabcolsep}{2.5pt}
\begin{tabular}{@{}lllllll@{}}
\toprule
\textbf{Domain} & \textbf{Split by} & \textbf{Train/val/test} & \textbf{Length} & \textbf{Rate} & \textbf{Warm-up} & \textbf{Seeds} \\ \midrule
FlareDW & episode & 458/106/141 & ${\sim}\SI{3}{s}$ & \SI{200}{Hz} & \SI{0.5}{s} & 7 \\
Ship convoy & episode & 34/6/12 & \SI{600}{s} & \SI{10}{Hz} & \SI{60}{s} & 7 \\
Fish & time & 65/15/20\% & \SI{289}{s} & \SI{30}{Hz} & \SI{2}{s} & 7 \\
2D open loop & episode & 31/7/10 & \SI{11}{s} & \SI{100}{Hz} & \SI{2}{s} & 3 \\
2D closed loop & episode & 31/7/3 & \SI{11}{s} & \SI{100}{Hz} & \SI{2}{s} & 7 \\
Free flight & time & 60/20/20\% & ${\sim}\SI{66}{s}$ & \SI{100}{Hz} & \SI{1}{s} & 3 \\
Gantry & traverse & 60/20/20\% & ${\sim}\SI{6}{s}$ & \SI{125}{Hz} & \SI{1}{s} & 3 \\
\bottomrule
\end{tabular}
\end{table}

\Cref{fig:teaser} summarizes the conclusions of the study: its bar plots compare, in each of the six domains, the memory-less state-of-the-art MLP baseline against the best memory-based architecture. We now analyze each domain in detail.

\subsection{Agile quadrotors interactions}
\label{subsec:flare-dw}
We begin with a dataset from prior work,
FlareDW \cite{kharitenko2025spatiotemporal}, which uses a custom high-fidelity CFD simulator to model interactions between two P600 quadrotors (\SI{3}{kg} mass, \SI{0.6}{m} rotor-to-rotor distance).
It covers agile, non-stationary downwash disturbances at velocities of \SIrange{0.5}{4.0}{m/s}, well beyond the low-speed regimes ($<\SI{0.5}{m/s}$) typically addressed in prior work.
Recorded at 200 Hz over 30 minutes of simulated flight, it has four maneuvers: Fly Below, Fly Above, Swapping, and Fast Swapping (up to \SI{8}{m/s} relative velocity), and incorporates absolute velocities, thrust vectors and relative states. The observation is the 11-dimensional encoding of~\cite{kharitenko2025spatiotemporal}, the state of the art for close-proximity downwash prediction, shown to outperform alternative parameterizations~\cite{li2023nonlinear, smith2023so, shi2020neural, shi2021neural}, so the memory gains here are measured on top of the strongest memoryless representation. The sufferer's own thrust is withheld because, across this dataset, it correlates too strongly with the downwash label.

\Cref{fig:bar_plots} shows each model's root mean squared error (RMSE) in predicting downwash force.
Every temporal model improves on the memoryless baseline, which is last of the eight at \SI{1.74}{N} against \SI{1.33}{N} for Reservoir Computing, a \SI{23}{\percent} reduction in error. RC (ESN) and Cross-Attention perform best, separated by less than their seed spread. Cross-Attention may perform well because it weights the source's past states around the time the wake was shed rather than committing to one of them. Delay Embedding does commit to a single delayed read and trails the other temporal models, possibly because the arriving downwash superposes several past emissions. History MLP and the sequential models (GRU, TCN, Mamba) are competitive, separated by little more than their seed spreads. We test the models open loop here because the dataset comes from prior work~\cite{kharitenko2025spatiotemporal} and we do not have access to the simulator. Open-loop accuracy may not translate to closed loop, where the observations are by nature out of distribution: the multirotor in the training data was not compensating for the downwash, so under active compensation the dip seen there no longer appears. We therefore also test closed-loop performance in some of the following domains.

\subsection{\texorpdfstring{Ship--ship wake in a shallow-water convoy}{Ship--ship wake in a shallow-water convoy}}

Next, we use a two-ship convoy simulator in which the wake is an explicit shallow-water wave field.
A sufferer trails a producer by \SIrange{250}{450}{m} in water $1.5\times$ its draught. The free surface obeys the linearized shallow-water equations, forced by the producer as a moving pressure patch, the standard device for generating ship-borne waves numerically~\cite{david2017shipwaves}. The solved field holds a depression locked under the producer and free long waves radiated whenever its speed or course changes, crossing the water at $\sqrt{gh}\approx\SI{11.3}{m/s}$, so the wash arriving at the sufferer was shed \SIrange{13}{35}{s} earlier. The label is the Froude--Krylov force of the incident field integrated along the sufferer's hull.
Its three outputs, the force along the separation, the force across it and the yaw moment, are each z-scored with their training statistics, and the RMSE is pooled over the three. 
Observations are the ship analog of the equivariant encoding of~\cite{kharitenko2025spatiotemporal}, in the separation-aligned frame.

All temporal models outperform the memoryless MLP (\Cref{fig:bar_plots}), and the best (TCN) cuts its error by 60\%.
\looseness=-1 The models that integrate the whole recent history (GRU, TCN, Cross-Attention, History MLP) lead, while Delay Embedding and RC (ESN) fall back toward the memoryless score. We attribute this to the wash at any instant superposing waves generated at many past times: the waves share one speed, but the producer's past track spans a range of distances, and distance sets the delay. On this data, a linear read at the exact arrival-matched delay recovers only \SI{60}{\percent} of the force variance the full window does, and bracketing it by two seconds more than closes the gap. The force follows the wave slope at the hull, a local phase one sample cannot carry and a small timing error inverts. By construction, history is the only route to the label: the recorded force is never applied to the sufferer's dynamics, and the producer's maneuvering decorrelates over \SIrange{10}{20}{s} while its wash takes \SIrange{13}{35}{s} to arrive.

\subsection{Fish Schooling}

Next, we use data from a robotic fish-schooling experiment designed to replicate a simplified two-fish subsystem~\cite{das2025fish}: an actively pitching NACA 0012 airfoil upstream and a compliant Mylar flag \SI{5.4}{cm} downstream.
We use only the single-pathway (hydrodynamic) condition, in which the flag responds passively, to isolate wake-mediated coupling. The data include camera tracking at \SI{60}{\hertz}, downsampled to \SI{30}{\hertz} for training, and laser Doppler velocimetry (LDV) recording streamwise flow velocity between the bodies.
The input features are airfoil pitching angle, LDV velocity, and flag front displacement; the target is flag tip displacement.
The cross-correlation between the airfoil's pitch angle and the flag-tip displacement peaks at a delay of ${\sim}\SI{0.33}{s}$. The disturbances are sinusoidal, with a nearly constant period and transport delay.

\begin{figure}[!b]
    \centering
    \includegraphics[width=0.385\columnwidth]{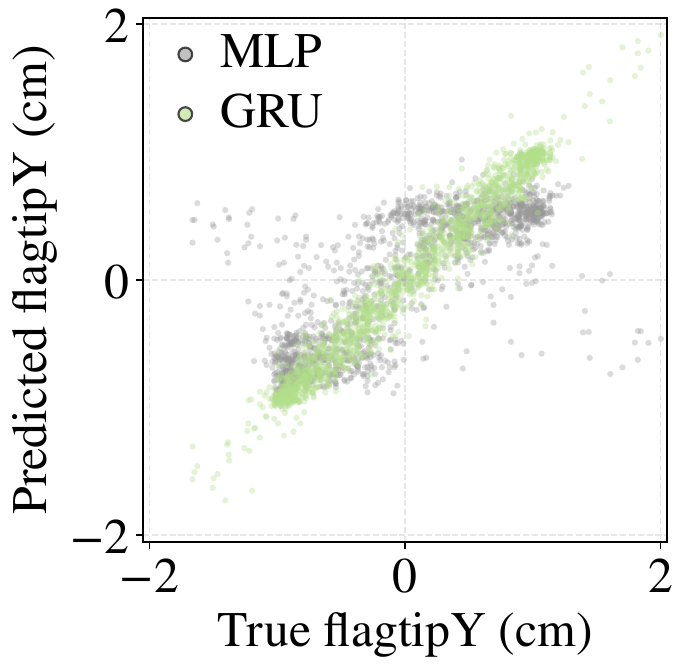}\hfill
    \includegraphics[width=0.60\columnwidth]{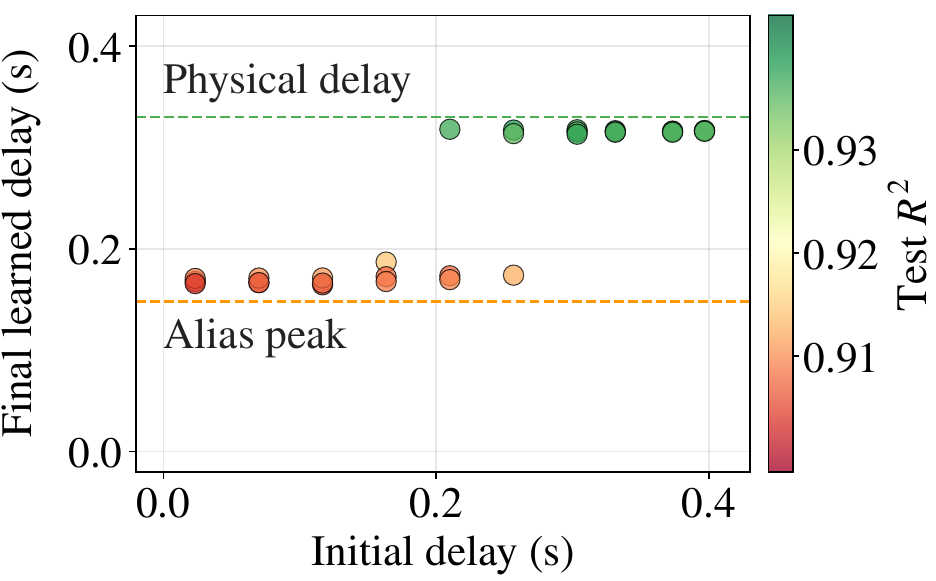}\\[2pt]
    \includegraphics[width=\columnwidth]{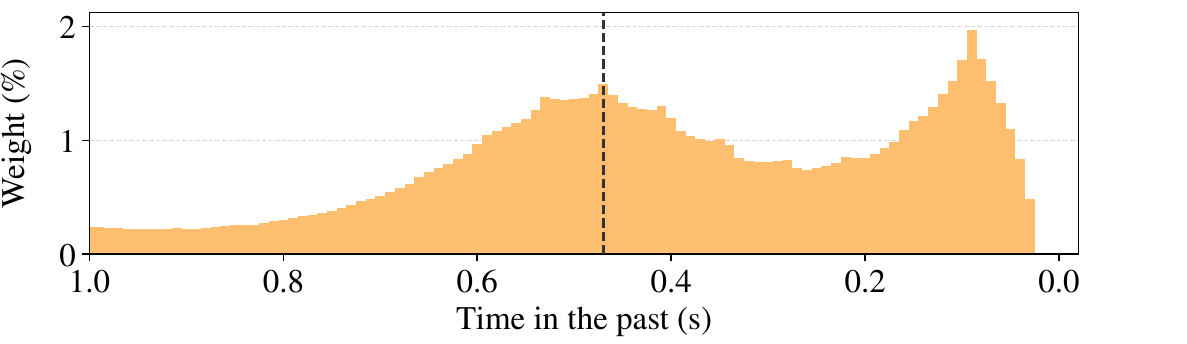}
    \caption{Fish schooling analysis (\textbf{top}), and, attention on the 2D downwash simulator (\textbf{bottom}). (\textbf{top left}) Scatter plot of true vs.\ predicted
    flag-tip displacement. The memoryless MLP ($R^2 = 0.57$) exhibits two distinct modes, failing to
    distinguish periodic phases, while the GRU achieves accurate, unbiased predictions.
    (\textbf{top right}) Convergence basin of the Delay Embedding model.
    When $\mu$ is randomly initialized, the model converges to either the true physical delay
    (0.33\,s, higher $R^2$) or an alias peak (0.148\,s, lower $R^2$).
    (\textbf{bottom}) Average attention weights of the Cross-Attention model over all episodes. The profile peaks at \SI{0.47}{s} (dashed line), a past lag consistent with the wake's emergent transit time.}
    \label{fig:fish}
\end{figure}

As in the prior two domains, models that incorporate temporal context perform best (\Cref{fig:bar_plots}). \Cref{fig:fish} also reports the coefficient of determination for the best (GRU) and the worst performer (baseline MLP).
The baseline MLP achieves a small RMSE, but with an $R^2$ of only $0.57$ and two correlation modes (clusters), which we attribute to two factors.
First, the wake vortexes take ${\approx}\SI{0.33}{s}$ to travel from the airfoil to the flag, so the MLP correlates the current foil angle with a flag response caused by a past state it cannot observe.
Second, the sinusoidal pitching introduces a phase ambiguity: the same foil angle occurs twice per cycle, once in ascent and again in descent, producing different wake structures each time.
History-aware models resolve these limitations by observing the foil angle over time, recovering the phase.
The GRU, with $R^2 = 0.96$, presents a more uniform spread about $y=x$, which indicates an unbiased modeling of the wake effect.

Delay Embedding trails the other models, and \Cref{fig:fish} shows a flaw specific to it: the delay it selects depends on the random initialization of its $\mu$ parameter. Because the disturbance is nearly sinusoidal, shifted copies of the signal at multiples of the half-period correlate almost as well as the true delay, creating alias correlation peaks that act as local minima. As a result, the model converges to the true physical delay only when $\mu$ is initialized in its basin of attraction; otherwise, it settles on an alias peak with lower $R^2$.

\subsection{2D \texorpdfstring{Navier--Stokes}{Navier-Stokes} Downwash Simulator}
We now introduce a custom 2D downwash simulation in which the multicopters are planar rigid bodies (XZ plane) that tilt and produce 2D thrust vectors.
The velocity field evolves by solving the incompressible Navier--Stokes equations with a fractional-step projection method~\cite{chorin1968numerical}, on a $200 \times 200$ grid spanning $\SI{20}{m} \times \SI{10}{m}$ at \SI{100}{Hz}. Each multirotor injects momentum through an actuator-disk source scaled by its instantaneous thrust~\cite{leishman2006principles}, so the wake column, its bending under motion, and its transport delay \emph{emerge} from the flow solution. To isolate the producer's wake from the sufferer's self-wash, the solver advances a second reference field containing only the latter, and the label is the drag difference between the two fields.
Each multicopter flies with a Linear Quadratic Regulator (LQR), and the predicted interaction forces enter as feedforward thrust and pitch corrections, so we can evaluate closed-loop tracking under three conditions: no compensation, an oracle with access to the true forces, and predictions from each trained model.
\looseness=-1 The observation is an 8D vector of relative states and the wake source's thrust components, with the sufferer's own thrust withheld for the same reason, and the output is the 2D disturbance force, all recorded at \SI{100}{\hertz}. Training separations span \SIrange{2.4}{3.4}{m}, and the test flights lie inside this range, at \SIrange{2.7}{3.2}{m}.

\begin{figure}[t]
    \centering
    \includegraphics[width=1.0\columnwidth]{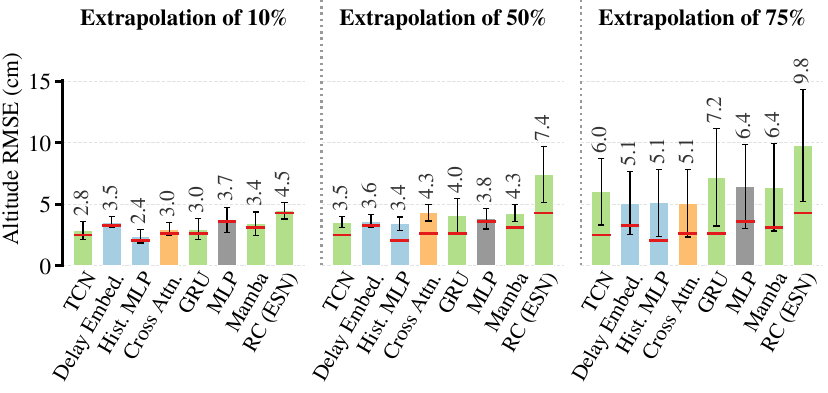}
    \caption{Out-of-distribution closed-loop ablation on the 2D downwash simulator.
    Models trained at the nominal physics are evaluated without retraining at $\pm 10\%$, $\pm 50\%$, and $\pm 75\%$ perturbation levels.
    Red lines denote the baseline closed-loop RMSE from a 3-seed evaluation at nominal conditions.
    Error bars show $\pm 1$ standard deviation over nine runs (3 seeds, 3 flights each).}
    \label{fig:ablation}
\end{figure}

\Cref{fig:bar_plots} shows the sufferer's tracking performance on random pre-planned trajectories, with every model tuned for the closed-loop altitude RMSE on 5 flights and scored on 3 separate flights, all unseen in training.
Models with temporal context again outperform the baseline MLP, except Reservoir Computing. It ranks last in closed-loop tracking, below the memoryless baseline, yet on the same simulator it is the second most accurate open-loop predictor of the force (\SI{1.04}{N} vs.\ \SI{1.01}{N} for the best and \SI{1.83}{N} for the baseline). Accurate open-loop prediction is therefore not sufficient on its own for closed-loop benefit.
We hypothesize that closing the loop is itself a mild distribution shift, and RC is the weakest in generalization (\Cref{fig:ablation}). The training data are recorded without compensation, so every episode carries the wake's signature in the sufferer's state (the altitude dip), and a model that compensates well removes the dip it learned from. RC's leading score on FlareDW (\Cref{subsec:flare-dw}), an open-loop test on data from the training distribution, is consistent with this: RC is accurate in-distribution but brittle outside it.

\looseness=-1 \Cref{fig:fish} (bottom) shows that the Cross-Attention weights, averaged over episodes unseen during training, peak at a time lag consistent with the emergent transit time of the jet  (dashed line). Similarly, in a targeted parameter sweep, the History MLP's error plateaus once its window is longer than the wake delay (see website). 
Mamba (16K) and Cross-Attention (18K) are smaller than the 65K baseline here, so their advantage is information, not capacity.

\begin{figure}[t]
    \centering
    \includegraphics[width=0.98\columnwidth]{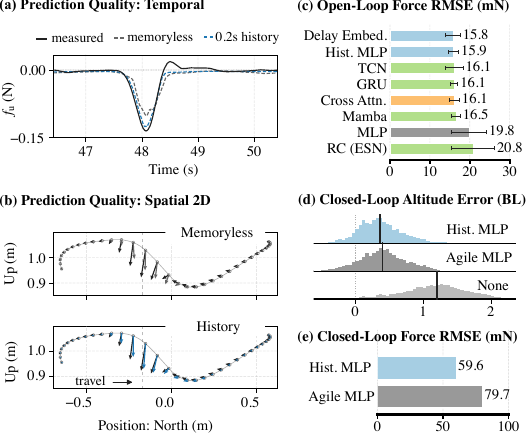}
    \caption{Free-flying quadrotor validations.
    \textbf{(a)} Force predictions over one traverse at \SI{5}{BL} vertical separation. \textbf{(b)} The same traverse in space, memoryless (top) and history (bottom). \textbf{(c)} RMSE for each model over 68 recordings.
    \textbf{(d)} Closed-loop altitude-error distribution over the nine settings of \Cref{tab:labdrone_cl} (lines mark means) and \textbf{(e)} mean in-flight prediction RMSE.
    In flight, the history model lowers altitude RMSE by \SI{65}{\percent} against no compensation and by \SI{10}{\percent} on average against the memoryless model. Error bars in (c) show $\pm 1$ standard deviation over 3 seeds.
    }
    \label{fig:labdrones}
\end{figure}

Finally, we use the simulator to test how well the different models \textbf{generalize} to fluid conditions out of the training distribution.
We take the models trained at the nominal physics and, without retraining, evaluate closed-loop RMSE with the actuator-disk strength, viscosity, and vertical separation drawn per episode within $\pm 10\%$, $\pm 50\%$, and $\pm 75\%$ of their nominal values. As shown in \Cref{fig:ablation}, every model degrades and the ordering holds exactly at $\pm 10\%$, but not beyond. At $\pm 50\%$ the memoryless baseline overtakes three temporal models, and at $\pm 75\%$ the leaders lie within their seed spread. RC (ESN) is last at every level and the only temporal model behind the baseline throughout, its error at $\pm 50\%$ nearly double the baseline's. What survives at every level is the top of the ranking, History MLP and TCN ahead of the baseline throughout, so the gains are not an artifact of overfitting the training distribution.

\subsection{Free-flying Quadrotors}\label{subsec:labdrones}

Our first hardware validation uses two lightweight (${\sim}\SI{70}{g}$, \SI{15}{cm} across the rotor ducts) brushless quadrotors~\cite{woo2026sanity} flying stacked in a motion-capture arena (\Cref{fig:teaser}e). The producer flies above the sufferer through varied patterns, at separations of \SIrange{0.5}{1.3}{m} and speeds of up to \SI{1.5}{m/s}. It reverses direction fast relative to the wake lag, so its position at emission time cannot be read from its current state. Both run the Freyja control stack~\cite{shankar2021freyja} on \SI{100}{Hz} OptiTrack data.

As in prior work~\cite{smith2023so}, we compute downwash as force residuals using the sufferer's orientation, commanded thrust and smoothed acceleration.
Observations use the same equivariant encoding as the other domains~\cite{kharitenko2025spatiotemporal}, with the sufferer's own thrust withheld as elsewhere.
Roughly $75$ minutes of usable open-loop flight are recorded across four scenario types---\emph{fixed-gap follow}, \emph{drifting-gap follow}, \emph{hovering sufferer} and \emph{crossing sweeps}---and each recording is cut chronologically 60/20/20\,\% into training, validation and test blocks (\Cref{tab:protocol}), a cut the three seeds rotate.

\Cref{fig:labdrones}a and c show force prediction accuracy and errors (RMSE) for the models.
Across all 68 recordings the memoryless MLP is second worst of the eight at \SI{19.8}{mN}, and every model with access to the past except RC (ESN) improves on it by \SIrange{17}{20}{\percent}.
Those six models span under \SI{5}{\percent} between themselves, inside their own seed spread.

We also incorporate the predictions as feed-forward terms in closed-loop feedback control, and evaluate tracking performance in nine configurations using no compensation, and using predictions from the memoryless and the history models.
On average, both models reduce altitude RMSE from \num{1.31} body-lengths (${\sim}\SI{19.6}{cm}$) to \num{0.51} BL (${\sim}\SI{7.6}{cm}$) and \num{0.46} BL (${\sim}\SI{6.9}{cm}$), respectively (\Cref{fig:labdrones}d), and the history model achieves a lower prediction RMSE in general.
\Cref{tab:labdrone_cl} reports the individual configurations: the history model outperforms in six of the settings, and is a close competitor in two additional \SI{1.8}{m/s} settings.
The results suggest that both models capture the majority of the forces, although the finer representation offered by the history model may play a bigger role in certain regimes with tighter coordination.

{
\definecolor{tabhdr}{rgb}{0.839,0.839,0.839}
\begin{table}[t]
\centering
\footnotesize
\caption{Closed-loop downwash compensation on the free-flying quadrotor pair.
Each cell is the sufferer's altitude RMSE over a \SI{65}{s} segment of a single flight, for \{\marker{catNoModel}\,no compensation \textbar{} \marker{catMemory}\,Agile MLP \textbar{} \marker{catHistory}\,History MLP\}, with the best in bold.
Separation and speed are the producer's commanded vertical gap and top speed, and both separation and RMSE are in body lengths ($1\,\text{BL}=\SI{15}{cm}$, the rotor-duct diameter).}
\label{tab:labdrone_cl}

\setlength{\tabcolsep}{3pt}
\setlength{\extrarowheight}{0pt}
\setlength{\aboverulesep}{0pt}
\setlength{\belowrulesep}{0pt}
\newcolumntype{Y}{>{$}c<{$}}
\newcolumntype{M}{>{\centering\arraybackslash $}m{1.75em}<{$}}
\newcommand{\fixedbox}[3]{
    \def\arraystretch{0.9}
    \setlength{\tabcolsep}{4pt}
    \begin{tabular}{M|M|M}
        #1 & #2 & #3
    \end{tabular}
}
\def\arraystretch{1.25}
\begin{tabular}{lYYYY} 
\toprule
        &       & \multicolumn{3}{c}{Speed (\si{m/s})}                 \\
        &       & {\cellcolor{tabhdr}}0.6 &{\cellcolor{tabhdr}}1.2 &{\cellcolor{tabhdr}}1.8  \\
 \multirow{3}{*}{\rotatebox{90}{Sep. (BL)}} {} & {\cellcolor{tabhdr}}3.3    & \fixedbox{1.49}{0.84}{\textbf{0.58}}  & \fixedbox{1.12}{\textbf{0.49}}{0.75}  & \fixedbox{1.44}{0.57}{\textbf{0.45}} \\
          & {\cellcolor{tabhdr}}4.7    & \fixedbox{1.11}{0.77}{\textbf{0.61}}  & \fixedbox{1.31}{0.42}{\textbf{0.36}}  & \fixedbox{1.15}{\textbf{0.31}}{0.35} \\
          & {\cellcolor{tabhdr}}6.0    & \fixedbox{1.60}{0.49}{\textbf{0.47}}  & \fixedbox{1.26}{0.45}{\textbf{0.30}}  & \fixedbox{1.27}{\textbf{0.25}}{0.27} \\
\bottomrule
\end{tabular}
\end{table}
}

\subsection{Monocopter Gantry}\label{subsec:gantry}

\begin{figure}[t]
    \centering
    \includegraphics[width=0.372\columnwidth,valign=t]{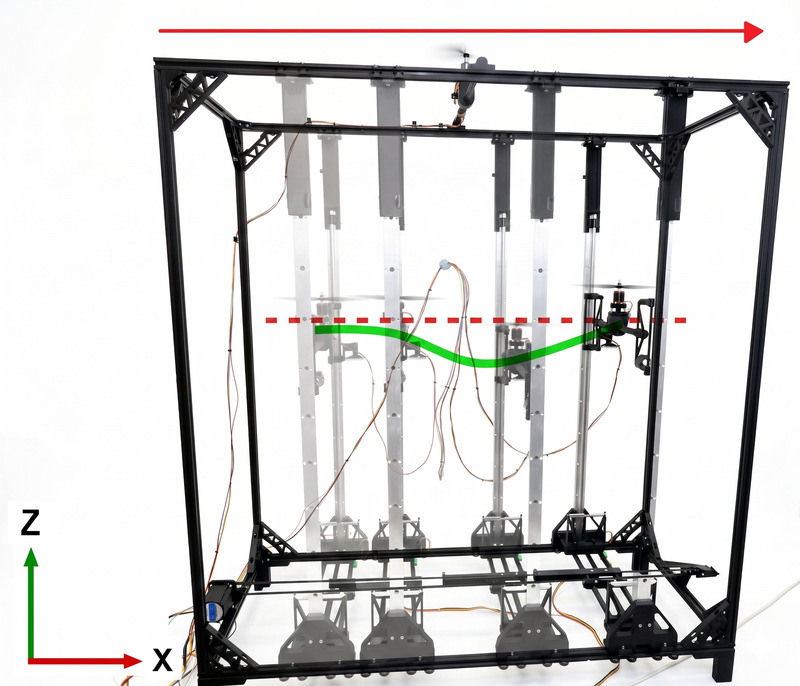}\hspace{0.02\columnwidth}\includegraphics[width=0.512\columnwidth,valign=t]{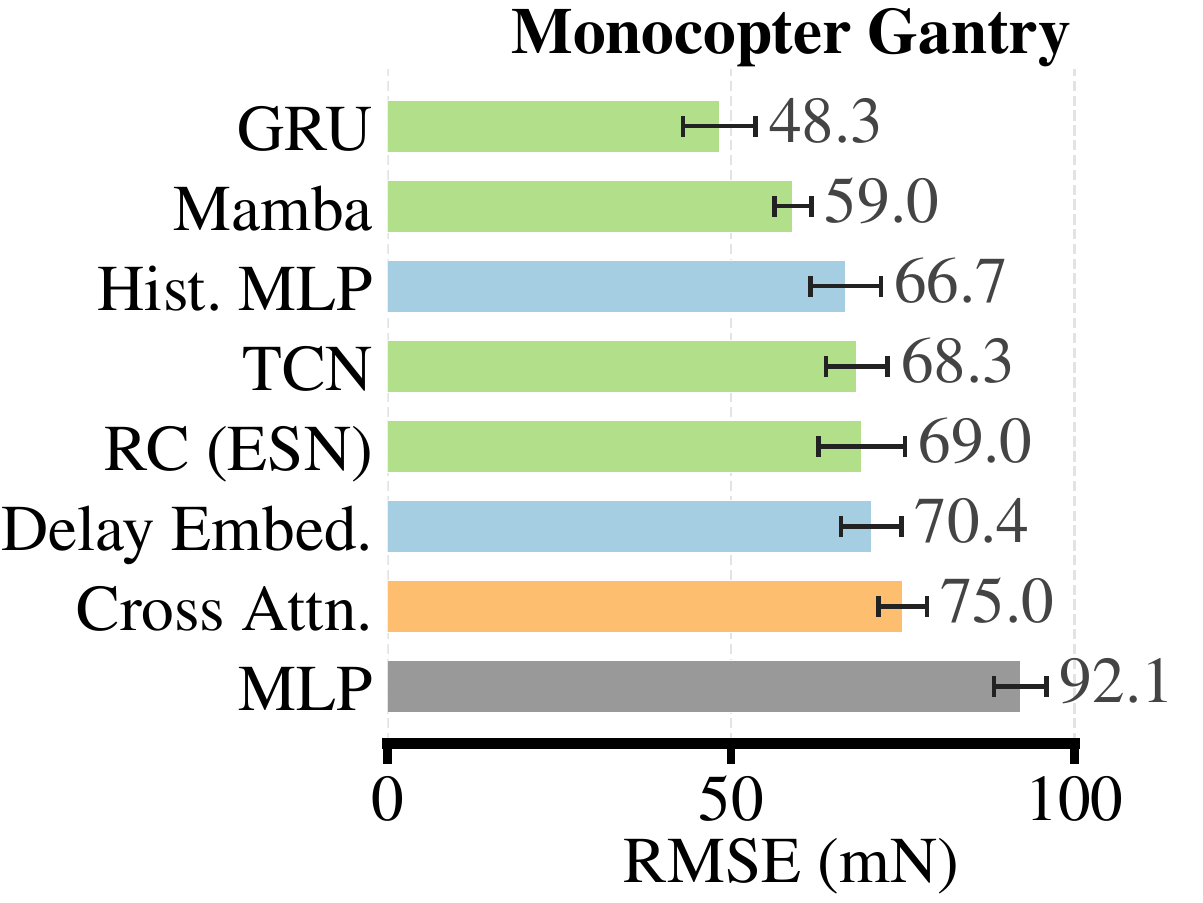}
    \caption{Constrained monocopter gantry.
    (\textbf{left}) One episode, overlaid. The source is fixed at the frame top and the sufferer, free to slide on vertical rails, is carried beneath it along the rail (red arrow). The green curve is the sufferer's measured altitude and the dashed line its target. 
    (\textbf{right}) Prediction RMSE of all architectures over the held-out traverses. The best model with memory (GRU) cuts the memoryless MLP's error by \SI{48}{\percent}. Error bars show $\pm 1$ standard deviation over 3 seeds.}
    \label{fig:gantry}
\end{figure}

As a final domain, we construct a custom monocopter gantry for accurate and repeatable experiments in controlled settings.
The system, shown in \Cref{fig:teaser}f and \Cref{fig:gantry}, uses two monocopters tethered to a rigid frame, for evaluations in the XZ plane.
The producer (source) is fixed in place on a tilt servo that can steer its wake jet in the 2D plane, and the sufferer slides freely on vertical rails and can be positioned along the x-axis by a belt-driven carriage.
Its 2D position is measured using range sensors at \SI{125}{Hz} read by an ATmega328P microcontroller, which also handles actuation.

We evaluate all models in a trial where the producer's thrust is randomized every \SI{0.1}{s} (faster than the wake transport delay), and the sufferer is commanded to hold one altitude, drawn per traverse from \SIrange{0.25}{0.45}{m}, while its carriage crosses the rail at a speed redrawn every \SI{0.15}{s}.
Over \num{161} traverses totaling \SI{17}{min}, the measured force on the sufferer correlates with the producer's commanded thrust at a \SI{304}{ms} lag.
In \Cref{fig:gantry}(right), every model beats the memoryless model in prediction RMSE, the best (GRU) by \SI{48}{\percent}.

\section{Discussion}\label{sec:discussion}

Our empirical results establish that temporal context consistently improves the modeling of fluid wake effects.
In all six domains the best temporal model outperforms the memoryless baseline, by \SIrange{20}{70}{\percent} depending on the medium---supporting our central thesis that the finite propagation delay of fluid wake entails modeling a temporal context.
However, the best temporal mechanism can vary with the medium.

These gains are consistent with wake transport. 
For instance, the Cross-Attention weights peak near the wake transport delay (\Cref{fig:fish}, bottom), and the Delay Embedding's favorable basin sits at the fish's measured \SI{0.33}{s} delay (top right).
On the gantry, the force correlates with the source's command at ${\sim}\SI{0.3}{s}$, not \SI{0}{s}.
The failure modes are also interpretable: aliasing under periodic forcing, and superposition that no fixed-lag measurement can capture.
Temporal context also carries derivatives and phase, and carries actuator dynamics and label filtering, which our data do not separate from transport, so we leave physical transport recovery as a hypothesis. 

\looseness=-1 \textbf{Model selection.} We recommend the History MLP as a reasonable default: it ranks about third of the eight models on average and first in closed loop on the 2D simulator, and on one CPU core it is typically only 1.5$\times$ slower than the memoryless MLP. The GRU (1.6$\times$ slower) is worth trying next. Delay Embedding is unreliable, coming first on free flight but last among the temporal models in three domains. The TCN (20$\times$ slower), Mamba (6$\times$) and Cross-Attention (4$\times$) are less suited to edge devices, although the TCN is worth its cost when long delays superpose, as on the ship convoy. RC (2$\times$ slower) holds up to 3.6M frozen weights in memory, and its closed-loop results (\Cref{fig:ablation}) make it a poor choice for control. 

A \textbf{limitation} we observe in this study is the absence of a clear winner among the models with temporal context, and the out-of-distribution ablation shows the ordering is trustworthy only near the physics a model was trained on. These conclusions are scoped to what the study measures: two robots, one wake source, and lab-scale speeds and separations. Within that scope, we draw two observations: \textit{(i)} support for the history of past states in the input accounts for the accuracy gains we measure, and \textit{(ii)} models that integrate the whole recent history read it most reliably across media.

\section{Conclusion}\label{sec:conclusion}

\looseness=-1 This work has studied the temporal properties a data-driven predictor needs to model fluid wake effects across CFD and experimental datasets, two custom simulators and two hardware platforms.
We establish that access to past states is a common requirement across the six domains, and that what the models learn is consistent with physical transport.
On both hardware platforms, most temporal models predicted the wake force better than the memoryless one. On the free-flying quadrotors, compensating with the history model reduced tracking error by about \SI{65}{\percent} against no compensation. Most of this comes from compensation itself (about \SI{61}{\percent} with the memoryless model). Against the memoryless model, the further reduction is about \SI{10}{\percent} on average across all nine settings. 

Future work will extend the study to more than two robots, where wake effects are more complex than the sum of pairwise interactions, and to predictors that adapt online.

\noindent\textbf{Code and data availability.} Code, the selected configuration of every model and domain, and our recorded or simulated data will be released at \url{https://wake-up-to-the-past.vercel.app/} on publication; FlareDW and fish data are from~\cite{kharitenko2025spatiotemporal,das2025fish}. 

\balance
\bibliographystyle{IEEEtran}
{\renewcommand{\footnotesize}{\fontsize{7.7}{8.8}\selectfont}\bibliography{references}}

\end{document}